\documentclass{svproc}
\usepackage{url}

\usepackage{graphicx}
\usepackage{float}
\usepackage{placeins}
\usepackage{multirow}
\usepackage{xcolor}
\usepackage{array}
\usepackage{colortbl}
\usepackage{makecell}
\usepackage{amsmath}
\usepackage{adjustbox}
\usepackage{amssymb}
\usepackage{algorithm}
\usepackage{algpseudocode}
\usepackage{wrapfig}
\usepackage{float}
\usepackage{placeins}
\usepackage{booktabs}

\begin{document}
\mainmatter              
\title{Parallel LLM Reasoning for Bias-Resilient, Robust Conceptual Abstraction}

\titlerunning{Contribution Title}  
%
\author{Aisvarya Adeseye\inst{1} \and Jouni Isoaho\inst{2} \and Adeyemi Adeseye\inst{3}}
\authorrunning{A. Adeseye et al} 
%
%
\institute{University of Turku, Turku, Finland,\\
\email{aisvarya.a.adeseye@utu.fi}\\ 
\and
University of Turku, Turku, Finland\\
\email{jouni.isoaho@utu.fi} \\
\and
Brilloconnetz Partners avoin yhtiö, Turku, Finland\\
\email{adeyemi@brilloconnetz.com}}

\maketitle              

\begin{abstract}



Large language models (LLMs) have been increasingly used to analyze text. However, they are often plagued with contextual reasoning limitations when analyzing long documents. When long documents are processed sequentially, early or dominant concepts can overshadow less visible but meaningful interpretations, leading to cumulative analytical bias, omission error, and over-generalization.
Additionally, independently generated outputs are often merged without systematic grounding, introducing redundancy, conceptual drift, and unsupported claims. This study proposes a structured framework combining parallel chunk-level processing with evidence-anchored consolidation. Texts are first divided into semantically coherent chunks and processed independently in parallel to remove influence from earlier processing. The independently generated interpretations are then consolidated using explicit evidence anchoring and prioritization that reduces dominance and over-generalization while improving traceability. Experiments with multiple model types and sizes indicate that parallel processing significantly reduces omission error by approximately 84\%, increases evidence traceability by up to 130\%, and reduces unsupported claims by up to 91\%. Smaller models benefited most, suggesting that efficient parallel chunking and consolidation play a critical role in achieving reliable and scalable textual analysis.

\end{abstract}
\section{Introduction}

Large language models (LLMs) have rapidly transformed large-scale textual interpretation, making automated thematic extraction \cite{maragheh2023llmtake}, conceptual abstraction \cite{opielka2025analogical}, qualitative coding \cite{adeseye2025llmassisted}, and long-document summarization \cite{godbole2025longcontext} possible in research and applied domains. Transformer-based architectures \cite{raaian2024review} combined with large-scale pretraining \cite{lin2024vila} have led to the creation of language models with demonstrated strong generative and reasoning capabilities \cite{liang2025pretraining}. 

These advances have encouraged the use of LLMs not merely as text generators, but as analytical instruments for synthesizing complex textual corpora \cite{sebastian2025generative}. However, reliable abstraction of long-form documents remains structurally constrained \cite{liu2024lost}. Most real-world corpora, like interview transcripts, policy reports, and multi-document datasets are too large to fit a single model context window; these are well-documented constraints of transformer architectures \cite{an2024makeyourllm,adeseye2025systematic}. Although larger models have shown improved contextual integration \cite{firstova2024contextual}, input length remains finite. Consequently, analysts must segment texts into smaller chunks prior to inference.

This segmentation into smaller chunks, if not properly implemented, introduces two recurrent structural failure modes. The first is \textit{cumulative analytical bias}. When text chunks are processed sequentially, each output influences the next one. Consequently, in autoregressive transformers \cite{dragunov2025sonar}, earlier interpretations shape later reasoning, reinforcing dominant themes while suppressing less visible but important ideas. Similar order-sensitivity and anchoring effects have been observed in iterative prompting and reasoning chains \cite{yadkori2024believe}. As a result, omission errors, reduced interpretive diversity, and positional dominance: concentrating attention towards the beginning and end of the text, but not on the mid-part of the text, a phenomenon known as lost in the middle, are problems that scaling alone does not fully resolve \cite{wan2025culturallens}. The second failure mode is \textit{ungrounded synthesis}. After chunk-level analysis, separate outputs must be combined into a single global structure. Without clear evidence constraints, this merging can create redundancy, conceptual drift, and unsupported claims. Consequently, hallucination and content not grounded in the input remains a known limitation of large language models \cite{adeseye2026hallucination}. In multi-stage pipelines, this risk increases as higher-level summaries may gradually move away from the original evidence. Prior research on faithful summarization and grounded generation emphasizes the need for evidence alignment and traceability \cite{ijctece2024stewardship,kambhamettu2025traceable}.

Importantly, these issues reveal that interpretive reliability is not solely a function of model capacity \cite{ioan2025interrater}. While scaling laws demonstrate systematic performance gains with parameter growth, scaling does not inherently eliminate order-dependence or enforce evidential grounding \cite{adeseye2025efficient}. Foundational models exhibit powerful emergent abilities \cite{lu2024emergent}, yet they remain sensitive to prompt structure, execution dynamics, and inference-time conditioning \cite{adeseye2025systematic,zagar2025dynamic,lu2025qinfer}. Thus, the methodological architecture of how inference is structured, constrained, and consolidated plays a key role in ensuring robustness, than parameter count alone. 

Therefore, this study develops and empirically validates a structured, bias-resilient framework for large-scale textual decomposition using LLMs by eliminating execution order dependence via parallel evidence chunk-level independent inference (PECII) by enforcing explicit evidence-anchored consolidation. Specifically, the study seeks to reduce omission error, dominance effects, and unsupported claims in long-form textual document analysis while improving traceability, and cross-model reliability, demonstrating that methodological inference design, rather than model scale alone, is a primary determinant of robust and scalable conceptual synthesis.

This paper makes four principal contributions:

\textbf{1. Formalization of Structural Bias in Sequential LLM Inference.}  
We model cumulative dominance, omission probability, and grounding feasibility within a constrained optimization framework, providing a theoretical account of order-induced bias in long-form analysis.

\textbf{2. Parallel Evidence-Constrained Independent Inference Architecture.}  
We introduce a boundary-aware, order-invariant parallel framework with explicit evidence validation and reliability-weighted consolidation to ensure robust textual inference.

\textbf{3. Multi-Dimensional Evaluation Framework.}  
We design an evaluation protocol assessing interpretive alignment, dominance mitigation, hallucination control, traceability, stability, scalability, and reproducibility.

\textbf{4. Empirical Demonstration of Structural Convergence Across Model Scales.}  
Parallel execution reduces omission error, increases traceability, and lowers unsupported claims, with model differences shrinking under structured execution.

Collectively, these findings show that reliable large-scale conceptual decomposition requires disciplined structural control over execution and evidence constraints.

\vspace{-0.2cm}
\section{Problem Formulation}

Large-scale textual interpretation with LLMs is structurally constrained. Because models have a finite context window, a full corpus cannot be processed in a single pass, forcing analysts to split it into smaller segments. This unavoidable division creates a multi-stage inference process in which outcomes depend on how segments are processed and later merged. Two recurring failures emerge. First, cumulative analytical bias arises in sequential processing, where early or dominant concepts act as implicit priors that shape and reinforce later interpretations, causing minor but meaningful themes to be overlooked. Second, ungrounded synthesis occurs when multiple chunk-level outputs are merged without strict evidence constraints. Although merging compresses content, it can also rewrite it, introducing unsupported claims, redundancy, blurred conceptual boundaries, and increased hallucination risk. Accordingly, we formalize these challenges as a constrained optimization problem.

We begin by defining the corpus. Let the transcript collection be:

{\scriptsize
\begin{equation}
\mathcal{D}=\{T^{(1)},T^{(2)},\dots,T^{(N)}\}.
\end{equation}
}

Each transcript is denoted as a sequence of atomic textual units like paragraphs or speaker turns:

{\scriptsize
\begin{equation}
T^{(n)}=(u^{(n)}_1,u^{(n)}_2,\dots,u^{(n)}_{m_n}).
\end{equation}
}

A chunking operator partitions each transcript into contiguous chunks with a token budget $L$:

{\scriptsize
\begin{equation}
\phi_L(T^{(n)}) \rightarrow \mathcal{C}^{(n)}=\{c^{(n)}_1,\dots,c^{(n)}_{k_n}\},
\qquad
\text{s.t. } \mathrm{tok}(c^{(n)}_j)\le L.
\end{equation}
}

The global chunk set is thus:

{\scriptsize
\begin{equation}
\mathcal{C}=\bigcup_{n=1}^{N}\mathcal{C}^{(n)},
\qquad
|\mathcal{C}| = M.
\end{equation}
}

Consequently, the analysis problem changes. We no longer read the full long-form textual document but a set of chunks. We therefore infer a set of abstract concepts from those chunks by defining the universe of possible valid concepts as $\Omega$. We also define the true but unknown set of concepts that are supported by the corpus as $\Omega^\star$. A concept is in $\Omega^\star$ if it is supported by at least 1 chunk. We represent support using an indicator function:

{\scriptsize
\begin{equation}
\mathsf{supp}(\omega;c)\in\{0,1\}
\end{equation}
}

where $\mathsf{supp}(\omega;c)=1$ means chunk $c$ contains sufficient evidence for concept $\omega$. The corpus-level support of $\omega$ is then:

{\scriptsize
\begin{equation}
\mathsf{Supp}(\omega;\mathcal{D})=
\mathbb{I}\left[\exists c\in\mathcal{C}:\mathsf{supp}(\omega;c)=1\right].
\end{equation}
}

The target concept set becomes:

{\scriptsize
\begin{equation}
\Omega^\star = \{\omega\in\Omega:\mathsf{Supp}(\omega;\mathcal{D})=1\}.
\end{equation}
}

Let $\mathcal{A}$ denote an inference procedure applied to $\mathcal{D}$, producing an estimated concept set $\widehat{\Omega} = \mathcal{A}(\mathcal{D})$. The central accuracy objective is that $\widehat{\Omega}$ should cover $\Omega^\star$.

Omission error measures how many truly supported concepts are missing in the extracted set:

{\scriptsize
\begin{equation}
\mathcal{E}_{\text{om}}(\widehat{\Omega})
=
1-\frac{|\widehat{\Omega}\cap \Omega^\star|}{|\Omega^\star|}
\end{equation}
}

We now formalize the second failure: ungrounded synthesis. Each extracted concept claim should be traceable to evidence. We model a claim instance as:

{\scriptsize
\begin{equation}
z=(\omega,e)
\end{equation}
}

Grounding feasibility is:

{\small
\begin{equation}
\mathcal{G}(\omega,e)=
\mathbb{I}[e\subseteq \mathrm{Trace}(c)]
\cdot
\mathbb{I}[|q(e)|\ge \lambda]
\cdot
\mathbb{I}[\mathcal{S}(\omega,e)\ge \tau]
\end{equation}
}

For an extracted set of instances $\mathcal{Z}$, we define:

{\scriptsize
\begin{equation}
\mathcal{E}_{\text{hall}}(\mathcal{Z})
=
\frac{1}{|\mathcal{Z}|}
\sum_{(\omega,e)\in\mathcal{Z}}
\left(1-\mathcal{G}(\omega,e)\right)
\end{equation}
}

The full theoretical problem can now be written as a single grand constrained objective:

{\small
\begin{equation}
\min_{\widehat{\Omega},\mathcal{Z},\mathcal{K}}
\Bigg[
\mathcal{E}_{\text{om}}(\widehat{\Omega})
+
\lambda_1 \mathcal{B}_{\text{order}}
+
\lambda_2 \mathcal{E}_{\text{hall}}(\mathcal{Z})
+
\lambda_3\big(\alpha_1\mathcal{R}(\mathcal{K})+\alpha_2\mathcal{L}(\mathcal{K})\big)
\Bigg]
\end{equation}
}

subject to:

{\small
\begin{equation}
\forall (\omega,e)\in\mathcal{Z}: \mathcal{G}(\omega,e)=1,
\qquad
\forall K_k\in\mathcal{K}: |D(K_k)|\ge \eta
\end{equation}
}

This formulation clarifies the contribution space. The problem is not only extracting patterns from text. The problem is controlling order-induced dominance, minimizing omission, enforcing evidence feasibility, and consolidating interpretations without redundancy and leakage. PECII is designed as a structural solution to this joint optimization problem.

\section{Parallel Evidence-Constrained Independent Inference (PECII)}

\begin{figure}
    \centering
    \includegraphics[width=0.6\linewidth]{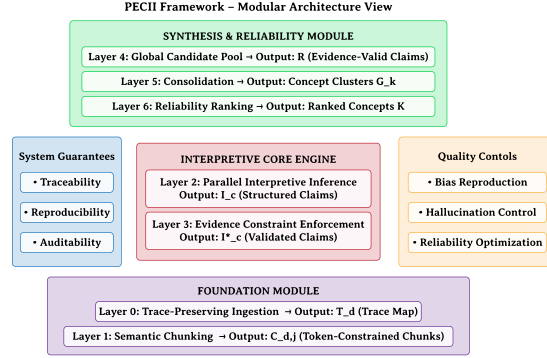}
    \caption{Parallel Evidence-Constrained Independent Inference (PECII)}
    \label{fig:PECII}
    \vspace{-0.6cm}
\end{figure}

PECII is a layered architecture for large-scale textual interpretation via LLM for long-form textual collections where the total content exceeds the context limits, targeting cumulative analytical bias and ungrounded synthesis (See Figure \ref{fig:PECII}. PECII introduces two structural ideas. First, independent inference through parallel chunk execution. Each chunk is processed without access to other chunk outputs, removing execution-order dependency. Second, evidence-constrained synthesis, with each interpretation requiring traceable evidence. Consolidation is then performed under explicit constraints on traceability, diversity, and redundancy. This design makes the output easier to audit and reproduce, improves interpretive coverage for smaller models, lowers unsupported claims, improves coherence and stability, while reducing omission error and dominance effects. 
It also scales efficiently through parallelism. The full pipeline is implemented end-to-end. Layers 0 and 1 below are implemented in Python. 
\vspace{-0.4cm}
\FloatBarrier
\vspace{-0.3cm}
\subsection{Layer 0: Trace-Preserving Long-Form Textual Document Ingestion and Normalization}

This layer converts each long-form textual document into traceable textual objects. The key output is not only the extracted text, but also the mapping to the source document. This mapping is required for evidence anchoring in later layers. It is also required for reproducibility and auditing. Let $d$ denote a document and $p$ a page. The extractor returns a set of spans:

\vspace{-0.15cm}
{\scriptsize
\begin{equation}
T_d =
\left\{
(s_{d,p,r}, \pi_{d,p,r})
\right\}_{p=1}^{P_d}
\vspace{-0.15cm}
\end{equation}
}
where
\vspace{-0.15cm}
{\scriptsize
\begin{equation}
\pi_{d,p,r} = (p, a_{d,p,r}, b_{d,p,r})
\vspace{-0.15cm}
\end{equation}
}
$s_{d,p,r}$ is the extracted string for region $r$, and $[a_{d,p,r}, b_{d,p,r}]$ are character offsets in the page text stream. We treat $\pi_{d,p,r}$ as the minimal trace key. It supports page-level retrieval and fine-grained span retrieval, and future quote validation. A normalization function $N(\cdot)$ reduces noise while preserving offset consistency. This removes recurring headers and footers, repairs hyphenation across line breaks, standardizes whitespace, and preserves speaker tags when present. The key constraint is that normalization must not destroy span references. If offsets become invalid, later evidence checks become unreliable. This is why Layer 0 is a trace-preserving transformation. It is not a free-form cleanup step. 
Layer 0 is critical because it defines what counts as evidence later. If extraction is noisy, later inference becomes noisy. If the trace is missing, later anchoring becomes unverifiable. If normalization is too aggressive, meaning can be lost. Therefore, PECII treats ingestion as a constrained preprocessing stage, designed to support evidence-grounded interpretation rather than only readability.

\vspace{-0.2cm}
\subsection{Layer 1: Semantic Chunking Under Token Constraints}

This Layer splits each long-form textual document into semantically coherent chunks. The chunks must fit within a token budget $L$. This layer is central for smaller models. Smaller models have shorter context and weaker long-range integration. They are more likely to skip minor interpretations in long-form texts. Chunking helps reduce the amount of text to reason over, which indirectly increases coverage. It also improves stability because each inference call sees a bounded context. Let the long-form text $T_d$ be a sequence of atomic units:

\vspace{-0.15cm}
{\scriptsize
\begin{equation}
T_d = \{u_1, \dots, u_m\}
\vspace{-0.15cm}
\end{equation}
}
Chunks are contiguous unions:

\vspace{-0.15cm}
{\scriptsize
\begin{equation}
C_{d,j} = \bigcup_{t=\ell_j}^{r_j} u_t
\vspace{-0.15cm}
\end{equation}
}
subject to

\vspace{-0.15cm}
{\scriptsize
\begin{equation}
\text{token Length}(C_{d,j}) \le L
\vspace{-0.15cm}
\end{equation}
}
PECII uses boundary-aware chunking, preferring to cut where topical continuity drops. It also avoids producing very small chunks. A boundary score estimates semantic shift between adjacent units:

\vspace{-0.15cm}
{\scriptsize
\begin{equation}
B(t) = 1 - \cos(\text{emb}(u_t), \text{emb}(u_{t+1}))
\vspace{-0.15cm}
\end{equation}
}
The cosine term uses unit embeddings. A larger $B(t)$ indicates a stronger boundary. In practice, the algorithm grows a chunk until it is close to $L$. It then checks for a strong boundary to cut chunks if the boundary is strong and the chunk is sufficiently large. This balances coherence, efficiency, and limits fragmentation.
This layer increases interpretive coverage, reduces overload, and makes the system scalable because chunks are the unit of parallel work. However, chunking alone does not remove cumulative bias. If chunks are processed sequentially, early outputs can still dominate later reasoning. This is why PECII treats chunking as a necessary preparation stage.
\vspace{-0.2cm}
\subsection{Layer 2: Parallel Interpretive Inference}
This performs interpretive inference on each chunk. The novelty is the independence constraint. Each chunk is processed in parallel and without access to other chunk outputs. This makes inference order-invariant. It reduces cumulative analytical bias and omission error because each chunk is treated as a first-class unit. Each chunk $c$ is processed independently:

\vspace{-0.15cm}
{\scriptsize
\begin{equation}
I_c = f_\theta(c; P)
\vspace{-0.15cm}
\end{equation}
}

The output is structured, contains claims, evidence, and trace pointers. Structured output is:

\vspace{-0.15cm}
{\scriptsize
\begin{equation}
I_c =
\left\{
(h_i, q_i, \rho_i, \kappa_i)
\right\}_{i=1}^{m_c}
\vspace{-0.15cm}
\end{equation}
}

$h_i$ is the interpretation claim. $q_i$ is the supporting quote. $\rho_i$ is the trace pointer to the long-form textual document span. $\kappa_i$ is a confidence score. The prompt template $P$ fixes the task and the schema. It reduces output drift and increases reproducibility. It also improves parsing reliability. The inference prompt requires that every claim must be grounded in a quote from the same chunk. This is enforced again in Layer 3.
Parallel inference is central to PECII. It breaks the sequential feedback loop, preventing early chunk dominance. It also increases throughput, ensuring practicality for large transcript sets. However, higher memory is needed due to increased concurrency overhead from parallel execution. PECII accepts this trade-off because the quality and efficiency benefits are large in the reported results.
\vspace{-0.2cm}
\subsection{Layer 3: Evidence Constraint Enforcement}
This layer filters chunk-level interpretations using explicit evidence constraints. It is the main hallucination control mechanism. LLMs may generate plausible but unsupported claims. This risk increases when outputs are later merged and rewritten. Therefore, PECII validates each claim before consolidation. Each claim $h$ must be supported by a quote $q$ that exists in the chunk trace map. Semantic support is computed as:

\vspace{-0.15cm}
{\scriptsize
\begin{equation}
\text{support}(h,q) =
\cos(\text{emb}(h), \text{emb}(q))
\vspace{-0.15cm}
\end{equation}
}
PECII uses a strict acceptance condition. The condition checks span validity, minimum quote lengths, and semantic alignment with a threshold $\tau$. A claim is accepted if:

\vspace{-0.15cm}
{\scriptsize
\begin{equation}
\mathbf{1}[\rho \in \text{Trace}(c)]
\cdot
\mathbf{1}[|q| \ge \lambda]
\cdot
\mathbf{1}[\text{support}(h,q) \ge \tau]
= 1
\vspace{-0.15cm}
\end{equation}
}
An optional verifier can be used for difficult cases. The verifier can be a SML check. The verifier returns a binary decision:

\vspace{-0.15cm}
{\scriptsize
\begin{equation}
g(h,q) \in \{0,1\}
\vspace{-0.15cm}
\end{equation}
}
The verifier is applied after the embedding filter, which reduces compute overhead and improves precision. It outputs the evidence-valid set $I_c^*$. This set is used for global aggregation. This is also the layer that drives large reductions in unsupported claim rate in the process. This layer is important because it converts interpretive inference into constrained inference, supporting traceability and auditability, while reducing speculative abstraction. It also improves consolidation quality by removing noisy claims before clustering. The main limitation is that strict filtering can remove rare but valid interpretations if the quote selection is poor. PECII mitigates this by requiring the model to return quotes explicitly and also by allowing multiple quotes per claim when needed.

\subsection{Layer 4: Global Candidate Pool}

This layer builds a global pool of all evidence-valid interpretations. It prepares the system for synthesis, preserves trace pointers, and confidence scores. It also creates embeddings for similarity-based grouping. Each valid interpretation becomes:

\vspace{-0.15cm}
{\scriptsize
\begin{equation}
r = (h, v_h, \rho, d, c, \kappa)
\vspace{-0.15cm}
\end{equation}
}
$h$ is the claim. $v_h$ is the claim embedding. $\rho$ is the trace pointer. $d$ is the document ID. $c$ is the chunk ID. $\kappa$ is confidence. The global pool is:

\vspace{-0.15cm}
{\scriptsize
\begin{equation}
R = \bigcup_c I_c^*
\vspace{-0.15cm}
\end{equation}
}
This pool is a key PECII artifact. It is a stable interface between inference and consolidation. It also supports analysis. Coverage can be measured from $d$. Distribution can be measured across $c$. Evidence diversity can be measured from $\rho$. This layer is also where scale becomes manageable. Instead of consolidating raw text, PECII consolidates a controlled set of evidence-valid claims. The main risk is pooling too many near-duplicates. PECII handles this in Layer 5 using clustering and redundancy constraints.

\subsection{Layer 5: Evidence-Constrained Consolidation}

This layer merges similar interpretations into consolidated concepts. The goal is to remove redundancy, preserve evidence traceability, and produce coherent boundaries. Consolidation begins with similarity grouping. PECII constructs a similarity graph over claim embeddings. The graph is:

\vspace{-0.15cm}
{\scriptsize
\begin{equation}
G = (V,E),
\quad
E = \{(i,j): \cos(v_i,v_j) \ge \delta\}
\vspace{-0.15cm}
\end{equation}
}
Nodes correspond to interpretations in $R$. Edges connect semantically similar claims. Clusters are then produced using connected components or community detection. The cluster set is:

\vspace{-0.15cm}
{\scriptsize
\begin{equation}
G = \{G_1, \dots, G_K\}
\vspace{-0.15cm}
\end{equation}
}
Each cluster is a candidate concept group. PECII then runs evidence-constrained consolidation per cluster. The consolidation prompt requires a definition, inclusion rules, and exclusion rules. Also, it needs a curated evidence list with trace pointers and a merge justification. This makes the synthesis auditable and also reduces over-generalization. PECII adds explicit constraints. It enforces evidence diversity across documents. Let $D_k$ be the set of distinct documents contributing to cluster $k$. The diversity constraint is:

\vspace{-0.15cm}
{\scriptsize
\begin{equation}
|D_k| \ge \eta
\vspace{-0.15cm}
\end{equation}
}
PECII also enforces evidence redundancy control. Evidence quotes must not be near-duplicates. Let $q_i$ be evidence quotes. The redundancy constraint is:
\vspace{-0.15cm}
{\scriptsize
\begin{equation}
\max_{i \ne j}
\cos(\text{emb}(q_i), \text{emb}(q_j))
\le \gamma
\vspace{-0.15cm}
\end{equation}
}
These constraints improve synthesis quality, reduce cross-concept leakage, and duplicate concepts. They also increase the justification quality of merges. This layer is where many pipelines fail because they merge outputs without constraints. PECII treats consolidation as a constrained optimization problem. It is implemented as a combination of algorithmic clustering and constrained LLM synthesis. The main trade-off is added compute for consolidation. PECII accepts this cost because consolidation is performed on clusters, not on all raw text.

\subsection{Layer 6: Reliability-Weighted Ranking}

This layer ranks consolidated concepts. It prioritizes concepts that are well supported and broadly represented. It also penalizes concepts that are unstable. This avoids simple frequency dominance. Let $D$ be the full document set. Let $D_K$ be the set of documents supporting concept $K$. Coverage is:

\vspace{-0.15cm}
{\scriptsize
\begin{equation}
\text{cov}(K) =
\frac{|D_K|}{|D|}
\vspace{-0.15cm}
\end{equation}
}
Evidence strength aggregates support between each claim and its evidence. Let $E_K$ be the evidence set for $K$. Evidence strength is:

\vspace{-0.15cm}
{\scriptsize
\begin{equation}
\text{estr}(K) =
\frac{1}{|E_K|}
\sum_{(h,q) \in E_K}
\cos(\text{emb}(h), \text{emb}(q))
\vspace{-0.15cm}
\end{equation}
}
Stability measures how consistent a concept is across runs or folds. Let $K^{(r)}$ be the concept instance extracted in run $r$. Let $J(\cdot,\cdot)$ be Jaccard overlap of evidence sets or keyword sets. Stability is:
\vspace{-0.15cm}
{\scriptsize
\begin{equation}
\text{stab}(K) =
\frac{2}{R(R-1)}
\sum_{r<s}
J(K^{(r)}, K^{(s)})
\vspace{-0.15cm}
\end{equation}
}
PECII combines these into a final score:
\vspace{-0.15cm}
{\scriptsize
\begin{equation}
\text{Score}(K) =
\alpha \cdot \text{cov}(K)
+
\beta \cdot \text{estr}(K)
+
\gamma \cdot \text{stab}(K)
\vspace{-0.15cm}
\end{equation}
}
Concepts are ranked by $\text{Score}(K)$. This ranking supports prioritization, reporting, and downstream analysis. It makes the final output more reliable, making it easier to justify why a concept is treated as important. The main sensitivity is a weighted selection. PECII treats $\alpha,\beta,\gamma$ as tunable parameters. It reports them for reproducibility. It also supports sensitivity analysis as part of evaluation.

\vspace{-0.2cm}
\subsection{Experimental Setting}

\begin{figure}
    \centering
\includegraphics[width=0.6\linewidth]{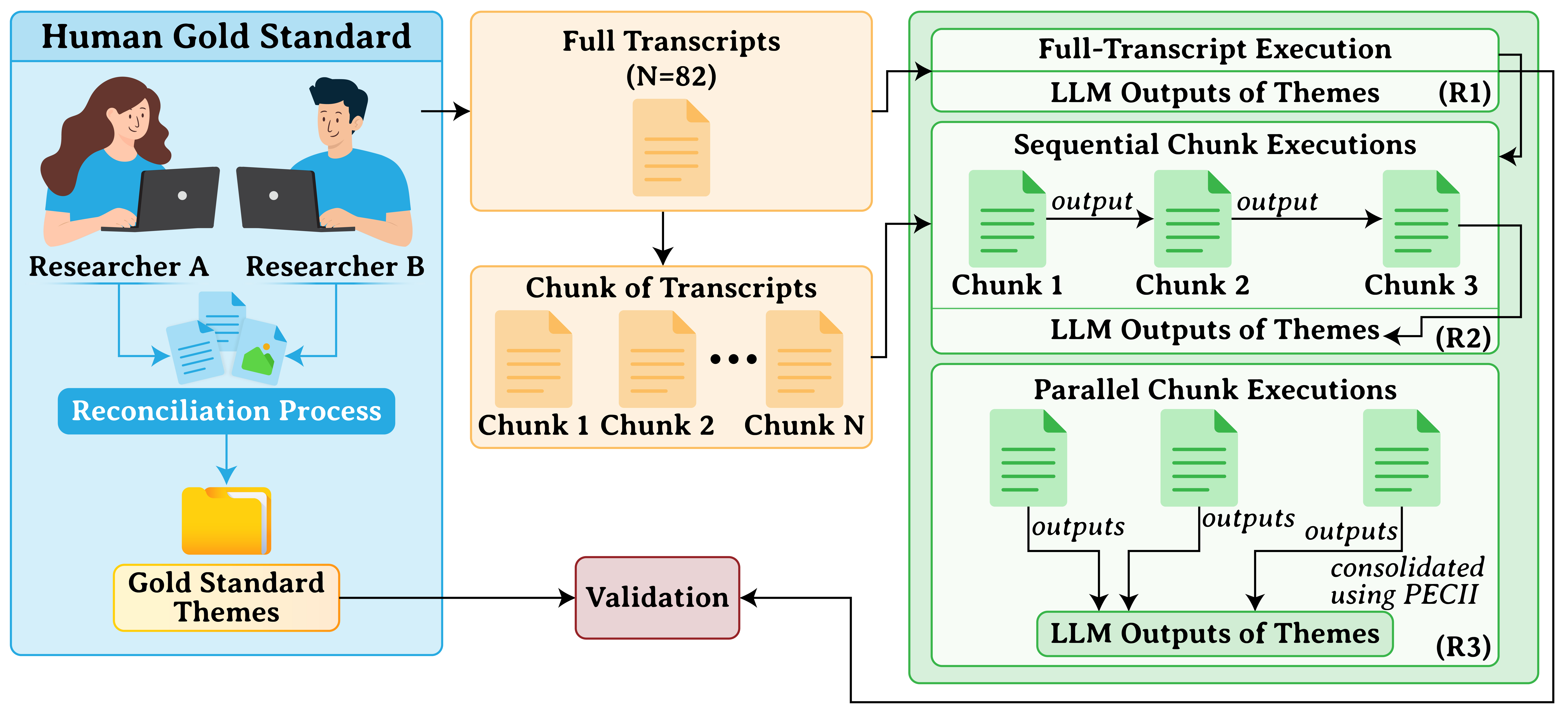}
    \caption{Experiment Setting of the Study}
    \label{fig:pecii}
    \vspace{-0.7cm}
\end{figure}

The experimental setting was designed to evaluate the robustness, bias mitigation capacity, and evidence-grounded reliability of the proposed PECII framework against a human-derived gold standard (See Figure \ref{fig:pecii}).

\vspace{-0.5cm}

\FloatBarrier

\textit{Human Ground Truth Construction:}
Two independent qualitative researchers conducted thematic analysis using NVivo-assisted coding. Each researcher independently coded the full interview transcripts and generated an initial set of themes, definitions, and supporting excerpts. The researchers were blinded to each other’s coding decisions during the initial phase to avoid mutual influence. After independent coding, both researchers’ theme sets were consolidated through a structured reconciliation process. Disagreements were identified at three levels: (1) theme presence or absence, (2) theme definition scope, and (3) excerpt-theme boundary assignments. For each disagreement, both researchers revisited the source transcript excerpts and provided justification for their interpretation. Where consensus could not be immediately reached, a re-examination of contextual segments was conducted, and alignment was resolved through discussion grounded strictly in textual evidence. The final reconciled theme set, including agreed definitions and validated supporting excerpts, was treated as the \textit{gold standard} reference set \( \Omega^* \) for subsequent quantitative validation.

\textit{LLM-Based Qualitative Analysis:}
The same interview transcripts were subsequently used as input for LLM-based qualitative analysis under three distinct execution conditions. First, in the \textbf{Direct Full-Transcript Execution} setting, the entire transcript was processed in a single inference call, subject to model token limitations, allowing the LLM to analyze the full contextual structure at once. Second, in the \textbf{Sequential Chunk Execution} approach, each transcript was divided into semantically coherent chunks constrained by a fixed token limit. These chunks were processed sequentially, where outputs from earlier chunks were incorporated into subsequent inference steps, thereby introducing cumulative contextual conditioning. Third, in the \textbf{Parallel Chunk Execution (PECII)} method, transcripts were segmented using boundary-aware chunking, and each chunk was processed independently and in parallel without access to outputs from other segments. This approach eliminated order-induced conditioning effects and reduced cross-chunk dependency biases, enabling a more controlled assessment of model behavior across isolated contextual units.


\textit{Consolidation Strategy:}
For chunk-based outputs, consolidation was performed via the proposed PECII evidence-constrained consolidation mechanism. Only themes that satisfied grounding feasibility criteria (traceable quotes and semantic alignment thresholds) were retained for final synthesis.

\textit{Validation Against Gold Standard:}
All model-generated theme sets from the three execution strategies were evaluated against the human-derived gold standard \( \Omega^* \). Validation metrics included omission error, early-chunk dominance, interpretive novelty, evidence traceability, unsupported claim rate, theme compression ratio, cross-theme leakage, and merge justification quality. 
Semantic alignment between model themes and gold standard themes was computed using cosine similarity over shared embedding space with a fixed matching Z-score. 
This experimental design enables direct comparison between human thematic synthesis and multiple LLM execution strategies, isolating the impact of execution order and consolidation structure while maintaining a consistent validation baseline.
\vspace{-0.2cm}
\subsection{LLM Models}

To evaluate the structural robustness and scale-invariance of the proposed PECII framework, six large language models were selected across different architectures and parameter sizes: LLaMA-1B, Qwen-1.5B, LLaMA-3B, Qwen-4B, LLaMA-8B, and ChatGPT 5.2. The selection followed three main principles: scale diversity, architectural diversity, and practical relevance. The models range from small (1B–3B) to mid-scale (4B–8B) and a large proprietary system (ChatGPT), allowing systematic testing of scaling effects under identical execution conditions. Smaller models were more sensitive to context overload and cumulative bias, while larger models were better able to capture and integrate complex semantic patterns across chunks. Including both LLaMA and Qwen ensures cross-family evaluation, reducing the risk that results depend on one specific architecture. ChatGPT serves as a high-capacity production baseline. A key research question is whether structured inference design reduces the performance gap between small and large models. Smaller models are efficient and suitable for resource-constrained environments but often struggle with long-form text. All models were tested using the same chunking strategy, similarity thresholds, evidence constraints, and consolidation steps, without model-specific prompt tuning. This controlled setup ensures that performance differences reflect model capacity and execution structure rather than prompt optimization. Overall, this selection enables analysis of parameter scaling, cross-architectural consistency, and whether methodological structure improves interpretive reliability independent of model size.

\vspace{-0.3cm}
\subsection{Data}

The data used in this study were collected through semi-structured interviews with 82 participants representing diverse organizational contexts, including non-governmental organizations (NGOs), private companies, universities, government institutions, and healthcare organizations. The primary objective of these interviews was to explore potential privacy concerns associated with introducing gamification in organizational settings, particularly in the context of employee surveys and workforce studies. Each interview lasted between 45 and 60 minutes, generating transcripts ranging from approximately 8,000 to 13,000 words. The semi-structured format allowed the researchers to use open-ended questions while maintaining thematic consistency across interviews. This approach enabled participants to express their perspectives in depth, capturing nuanced opinions, contextual experiences, and potential risks related to privacy and data handling. All participants provided explicit informed consent prior to participation. Ethical procedures were followed throughout the study. The transcripts were fully anonymized before analysis, with all personally identifiable information removed. Anonymization was independently verified by both researchers to ensure consistency and data integrity before conducting the qualitative coding and LLM-based analysis.

\vspace{-0.2cm}
\subsection{Validation Matrix }

\paragraph{Omission Error Rate (\%):}
This measures the proportion of expert-supported concepts not extracted by the model:
\vspace{-0.15cm}
{\scriptsize
\begin{equation}
E_{om} = \left( 1 - \frac{|\hat{\Omega} \cap \Omega^{*}|}{|\Omega^{*}|} \right) \times 100
\vspace{-0.15cm}
\end{equation}
}
$\Omega^{*}$ represents an expert reference concept set, $\hat{\Omega}$ represents the extracting model concept set

\paragraph{Early-Chunk Dominance Index (ECDI):}
Quantifies the proportion of extracted concepts originating from early chunks under sequential execution:
\vspace{-0.15cm}
{\scriptsize
\begin{equation}
\mathrm{ECDI} =
\frac{\sum_{c \in C_{\text{early}}} f(c)}
{\sum_{c \in C} f(c)}
\vspace{-0.15cm}
\end{equation}
}
It detects order-induced anchoring bias from autoregressive conditioning.  
Replication requires labeling chunks by order, defining an early window (e.g., first 20\%), counting extracted concepts from early chunks, and dividing by total extracted concepts.

\paragraph{Interpretive Novelty Score (1--5):}
Evaluates analytical depth beyond surface paraphrasing:
\vspace{-0.15cm}
{\scriptsize
\begin{equation}
\mathrm{INS} =
\frac{1}{|\Theta|}
\sum_{\theta \in \Theta} n(\theta)
\vspace{-0.15cm}
\end{equation}
}
This ensures diversity and conceptual richness. Replication requires blind expert rating of each theme on a 1-5 scale and averaging across themes and raters.

\paragraph{Evidence Traceability Score (\%):}
Calculates the proportion of themes supported by sufficient multi-source evidence:
\vspace{-0.15cm}
{\scriptsize
\begin{equation}
\mathrm{ETS} =
\frac{
\left|
\left\{
\theta :
|Q(\theta)| \geq k \ \wedge \
|S(\theta)| \geq m
\right\}
\right|
}
{|\Theta|}
\times 100
\vspace{-0.15cm}
\end{equation}
}
This is the primary grounding and auditability metric.  
Replication requires verifying that each theme contains at least $k$ quotes from at least $m$ distinct transcripts.

\paragraph{Unsupported Claim Rate (\%):}
Measures hallucination frequency:
\vspace{-0.15cm}
{\scriptsize
\begin{equation}
\mathrm{UCR} =
\frac{\text{Unsupported Claims}}
{\text{Total Claims}}
\times 100
\end{equation}
}
It directly captures grounding failure.  
Replication requires verifying that each claim has a valid trace pointer and semantic alignment with supporting quotes (cosine similarity $\geq 0.75$).

\paragraph{Theme Compression Ratio (TCR):}
Quantifies redundancy reduction after consolidation:
\vspace{-0.15cm}
{\scriptsize
\begin{equation}
\mathrm{TCR} =
\frac{|U|}
{|\Theta|}
\vspace{-0.15cm}
\end{equation}
}
It evaluates abstraction efficiency.  
Replication requires counting raw candidate claims prior to consolidation and dividing by final consolidated themes.

\paragraph{Cross-Theme Leakage (\%):}
Measures conceptual boundary violation:
\vspace{-0.15cm}
{\scriptsize
\begin{equation}
\mathrm{CTL} =
\frac{
\left|
\left\{
u \in U :
u \text{ assigned to multiple themes}
\right\}
\right|
}
{|U|}
\times 100
\vspace{-0.15cm}
\end{equation}
}
It detects poor clustering or merge overreach.  
Replication requires tracking excerpt-to-theme assignments and computing the proportion appearing in multiple themes.

\paragraph{Merge Justification Quality (1--5):}
Evaluates the clarity and strength of consolidation rationale:
\vspace{-0.15cm}
{\scriptsize
\begin{equation}
\mathrm{MJQ} =
\frac{1}{|\Theta|}
\sum_{\theta \in \Theta} j(\theta)
\vspace{-0.15cm}
\end{equation}
}
This ensures merges are principled rather than arbitrary compression. It requires blind expert evaluation of merge explanations and average scores across themes.

\vspace{-0.3cm}
\section{Performance Validation}

\subsection{Agreement and Reliability Validation}

Before evaluating execution strategies and consolidation effects, we first validated the reliability of the human-derived gold standard and quantified agreement between model outputs and expert thematic analysis. Two independent researchers conducted NVivo-assisted coding on the full transcript corpus. Inter-rater reliability reached $\kappa = 0.84$ for theme presence, $\kappa = 0.79$ for excerpt–theme assignment, and $\kappa = 0.82$ for boundary inclusion decisions, indicating substantial to near-perfect agreement and confirming the statistical robustness of the consolidated gold standard. Disagreements (8.7\% of coded instances) were resolved through structured reconciliation grounded in transcript evidence. Model–gold agreement was then evaluated using Cohen’s $\kappa$, macro-averaged F1 score, and Jaccard similarity. At baseline (no chunking), average agreement across models yielded $\kappa = 0.46$, $F1 = 0.51$, and Jaccard $= 0.43$. Sequential chunking improved agreement to $\kappa = 0.63$, $F1 = 0.68$, and Jaccard $= 0.61$, while parallel chunk-level execution further increased agreement to $\kappa = 0.88$, $F1 = 0.90$, and Jaccard $= 0.89$. Cross-model variance in agreement decreased from 0.18 at baseline to 0.04 under parallel execution, indicating convergence toward a shared reliability ceiling. Stability analysis across five independent runs per configuration showed mean jaccard consistency improving from 0.58 (baseline) to 0.72 (sequential) and 0.91 (parallel). Paired Wilcoxon signed-rank tests confirmed that improvements across execution strategies were statistically significant ($p < 0.01$) with large effect sizes ($d = 1.12$–$2.47$). These agreement results establish that methodological structure, particularly parallel independence and evidence anchoring, substantially increases alignment with expert interpretation while reducing stochastic variability across models.

\vspace{-0.4cm}

\subsection{Validation of Cumulative Bias Mitigation through Parallel Chunk-Level Processing}

Tables \ref{tab:ea_its_bias_dominance_novelty} evaluate how execution strategy affects cumulative analytical bias and interpretive quality. We compared a no-chunking interview transcript baseline with chunked transcript processing under sequential and parallel execution strategies.From Table \ref{tab:ea_its_bias_dominance_novelty}, Omission error decreased consistently across models when chunking was introduced, indicating that transcript segmentation improves coverage. At baseline (no chunking), omission declines as model size increases (36.8\% in LLaMA-1B vs 22.8\% in ChatGPT), indicating that larger models handle long serial contexts better. This is consistent with transformer scaling principles: larger models maintain richer token-level representations, are less susceptible to attention dilution, and also tolerate long autoregressive chains better, which reduces context degradation across extended inputs.

Sequential chunk execution reduces omission by 36–40\% compared to the baseline, suggesting that dividing the transcript helps limit attention decay. Shorter segments reduce attention competition within each chunk.
However, parallel chunk execution led to approximately 84\% reduction compared to baseline across all models, indicating that omission is driven more by cumulative context-window saturation in sequential processing than by model size. In sequential execution, each interpretation becomes part of the next input, creating conditioning drift. Parallel execution removes this carryover effect by isolating chunks and processing them separately. Importantly, the performance gap across models decreases from 14\% at baseline to 2.2\% under parallel execution (\textgreater 80\% reduction in cross-model variance). This convergence shows that architectural scale mainly compensates for sequential bias accumulation rather than eliminating it. When sequential dependency is removed, models behave much more similarly.

A similar trend is observed for Early-Chunk Dominance, where baseline dominance decreases as model size increases (0.51 $\rightarrow$ 0.36). However, across all model sizes and types, sequential processing reduces dominance by approximately 38\%, whereas parallel processing achieves a substantially larger reduction of around 82\%. This pattern reflects positional bias in autoregressive transformers, where earlier tokens anchor later reasoning through repeated conditioning. Parallel execution resets the conditioning state, interrupting this reinforcement loop. Also, for Interpretive Novelty, sequential chunking produces modest gains (9.5–23.3\%), while parallel execution yields larger improvements (up to 56.7\%), especially for smaller models. Smaller models show greater improvement under parallel execution because they tend to lose thematic diversity during long sequential executions, whereas chunk isolation prevents this progressive narrowing of interpretation. In contrast, larger models exhibit smaller gains because they already preserve broader semantic coverage during sequential processing, leaving less room for additional improvement under parallel execution.

Overall, the execution strategy functions as a bias-mitigation mechanism. Parallel chunking acts as an inference-time regularizer, limiting cumulative interpretive drift and reducing cross-model disparities. This indicates that bias accumulation is not solely determined by model architecture, but also emerges from the sequential execution process itself, where earlier outputs progressively constrain subsequent reasoning.

\begin{table*}[htbp]
\centering
\scriptsize
\vspace{-0.5cm}
\caption{Cumulative Bias, Dominance, and Interpretive Novelty under Different Execution Strategies (Baseline: No Chunking)}
\label{tab:ea_its_bias_dominance_novelty}
\renewcommand{\arraystretch}{1.2}
\setlength{\tabcolsep}{0.8pt}

\newcommand{\chgcellbaseline}[5]{%
\begingroup
\setlength{\fboxsep}{0.9pt}
\setlength{\fboxrule}{0.5pt}
\begin{tabular}[c]{@{}c@{}}%
{\scriptsize
\fcolorbox{orange!60!black}{orange!15}{\textbf{#4}}%
\hspace{3pt}%
\fcolorbox{green!60!black}{green!15}{\textbf{#5}}}\\[-0.5pt]
\textcolor{blue}{#1} $\rightarrow$ \textcolor{orange}{#2} / \textcolor{red}{#3}
\end{tabular}%
\endgroup
}

\begin{tabular}{|l|c|c|c|}
\hline
\textbf{Model} &
\textbf{\makecell{Omission \\ Error Rate (\%)}} &
\textbf{\makecell{Early-Chunk \\ Dominance Index}} &
\textbf{\makecell{Interpretive \\ Novelty (1--5)}} \\
\hline

LLaMA-1B &
\chgcellbaseline{36.8}{23.5}{5.9}{-36.1\%}{-84.0\%} &
\chgcellbaseline{0.51}{0.32}{0.09}{-37.3\%}{-82.4\%} &
\chgcellbaseline{3.1}{3.8}{4.7}{+22.6\%}{+51.6\%} \\
\hline

Qwen-1.5B &
\chgcellbaseline{35.2}{22.1}{5.4}{-37.2\%}{-84.7\%} &
\chgcellbaseline{0.49}{0.30}{0.09}{-38.8\%}{-81.6\%} &
\chgcellbaseline{3.0}{3.7}{4.7}{+23.3\%}{+56.7\%} \\
\hline

LLaMA-3B &
\chgcellbaseline{30.6}{19.2}{4.8}{-37.3\%}{-84.3\%} &
\chgcellbaseline{0.44}{0.27}{0.08}{-38.6\%}{-81.8\%} &
\chgcellbaseline{3.4}{4.1}{4.8}{+20.6\%}{+41.2\%} \\
\hline

Qwen-4B &
\chgcellbaseline{27.4}{16.8}{4.3}{-38.7\%}{-84.3\%} &
\chgcellbaseline{0.41}{0.25}{0.08}{-39.0\%}{-80.5\%} &
\chgcellbaseline{3.7}{4.3}{4.8}{+16.2\%}{+29.7\%} \\
\hline

LLaMA-8B &
\chgcellbaseline{24.1}{14.9}{3.9}{-38.2\%}{-83.8\%} &
\chgcellbaseline{0.38}{0.23}{0.07}{-39.5\%}{-81.6\%} &
\chgcellbaseline{4.0}{4.5}{4.9}{+12.5\%}{+22.5\%} \\
\hline

ChatGPT &
\chgcellbaseline{22.8}{13.6}{3.7}{-40.4\%}{-83.8\%} &
\chgcellbaseline{0.36}{0.22}{0.07}{-38.9\%}{-80.6\%} &
\chgcellbaseline{4.2}{4.6}{4.8}{+9.5\%}{+14.3\%} \\
\hline

\end{tabular}
\vspace{-0.8cm}
\end{table*}

\FloatBarrier

\vspace{-0.2cm}

\subsection{Evaluation of Evidence-Anchored Consolidation}

Table \ref{tab:ea_its_evidence_consolidated} evaluates the impact of evidence-anchored consolidation on model performance.
Evidence traceability increased substantially across all architectures. LLaMA-1B improved from 41\% to 95\% (+131.7\%, +54 points), Qwen-1.5B from 44\% to 96\% (+118.2\%), and LLaMA-8B from 71\% to 97\% (+36.6\%). At baseline, traceability scales with model size (41–71\%). After anchoring, performance converges tightly to 95–97\%, compressing the cross-model spread from 30 points to 2 points ($\mathrel{>}$90\% reduction). This sharp variance collapse indicates structural convergence rather than incremental improvement. Absolute gains decrease with model size (54 $\rightarrow$ 52 $\rightarrow$ 38 $\rightarrow$ 31 $\rightarrow$ 26 points), demonstrating a diminishing-return gradient in which anchoring corrects instability in smaller models while refining already stable larger models.

Unsupported claims declined dramatically across all models (for example, LLaMA-1B: 34 $\rightarrow$ 3, -91.2\%; ChatGPT: 9 → 2, -77.8\%). The baseline spread (25 points) collapses to a 1-point band (2-3\%). This near-elimination of cross-model variance suggests that explicit evidence linkage constrains speculative generation and narrows the generative search space. All models converge to a common low-error regime. Theme compression increased consistently (absolute gains of 6–8 units), with final values clustering at 14.8–15.6. Cross-theme leakage simultaneously decreases by 66–79\%, and merge justification quality converges to 4.8–4.9 (near scale maximum). Importantly, compression improves while leakage and unsupported claims decrease, resolving the typical consolidation trade-off where abstraction increases error. Justification variance compresses from a 1.3-point spread to 0.1, indicating near-saturation of explanatory quality.

\vspace{-0.3cm}
\begin{table*}[htbp]
\centering
\scriptsize
\caption{Evidence Anchoring and Consolidation Effectiveness Across Models}
\label{tab:ea_its_evidence_consolidated}
\renewcommand{\arraystretch}{1.3}
\setlength{\tabcolsep}{1.0pt}

\newcommand{\chgcell}[3]{%
\begin{tabular}[c]{@{}c@{}}%
{\scriptsize
\fcolorbox{green!60!black}{green!15}{$\uparrow$ \textbf{#3}}}\\[-1pt]
\textcolor{blue}{#1} $\rightarrow$ \textcolor{red}{#2}
\end{tabular}%
}

\begin{tabular}{|l|c|c|c|c|c|}
\hline
\textbf{Model} &
\textbf{\makecell{Evidence \\ Traceability\\ (\%)}} &
\textbf{\makecell{Unsupported \\ Claim Rate \\ (\%)}} &
\textbf{\makecell{Theme \\ Compression \\ Ratio}} &
\textbf{\makecell{Cross-Theme \\ Leakage \\ (\%)}} &
\textbf{\makecell{Merge \\ Justification \\ Quality(1-5)}} \\
\hline

LLaMA-1B &
\chgcell{41}{95}{+131.7\%} &
\chgcell{34}{3}{-91.2\%} &
\chgcell{6.2}{14.8}{+138.7\%} &
\chgcell{29}{6}{-79.3\%} &
\chgcell{2.8}{4.8}{+71.4\%} \\
\hline

Qwen-1.5B &
\chgcell{44}{96}{+118.2\%} &
\chgcell{31}{3}{-90.3\%} &
\chgcell{6.5}{15.0}{+130.8\%} &
\chgcell{27}{6}{-77.8\%} &
\chgcell{2.9}{4.8}{+65.5\%} \\
\hline

LLaMA-3B &
\chgcell{58}{96}{+65.5\%} &
\chgcell{19}{2}{-89.5\%} &
\chgcell{7.4}{15.2}{+105.4\%} &
\chgcell{22}{5}{-77.3\%} &
\chgcell{3.3}{4.9}{+48.5\%} \\
\hline

Qwen-4B &
\chgcell{66}{97}{+47.0\%} &
\chgcell{14}{2}{-85.7\%} &
\chgcell{8.1}{15.4}{+90.1\%} &
\chgcell{18}{5}{-72.2\%} &
\chgcell{3.7}{4.9}{+32.4\%} \\
\hline

LLaMA-8B &
\chgcell{71}{97}{+36.6\%} &
\chgcell{11}{2}{-81.8\%} &
\chgcell{8.4}{15.6}{+85.7\%} &
\chgcell{16}{4}{-75.0\%} &
\chgcell{4.0}{4.9}{+22.5\%} \\
\hline

ChatGPT &
\chgcell{63}{96}{+52.4\%} &
\chgcell{9}{2}{-77.8\%} &
\chgcell{8.6}{15.3}{+77.9\%} &
\chgcell{15}{5}{-66.7\%} &
\chgcell{4.1}{4.8}{+17.1\%} \\
\hline

\end{tabular}
\vspace{-0.6cm}
\end{table*}

\FloatBarrier

Across all five metrics, traceability, unsupported claims, compression, leakage, and justification quality, post-anchoring results cluster within narrow bands. No metric deteriorates. The joint improvement across independent dimensions indicates coordinated constraint tightening rather than isolated metric optimization. Conceptually, evidence anchoring introduces an external constraint layer that requires explicit justification for each interpretation. This reduces speculative drift and limits scale-dependent variability. Anchoring therefore operates as a structural regularization mechanism: it equalizes capacity differences, reduces architectural variance, and aligns models toward a shared reliability ceiling. 
Overall, evidence-anchored consolidation does not merely enhance performance; it reshapes the inference structure itself. Reliability gains arise from disciplined constraint imposition rather than increased model capacity, producing stable, traceable, and coherent thematic synthesis across architectures.

\vspace{-0.3cm}
\section{Discussion}

The results demonstrate that methodological structure, specifically chunking, parallel execution, and evidence-anchored consolidation, plays a more critical role than model size alone in achieving reliable large-scale textual abstraction. Chunking provides the foundational improvement, particularly for smaller models. When transcripts are processed monolithically, smaller models exhibit higher omission error and dominance effects due to interpretive overload. Introducing chunk-level segmentation reduces cognitive burden per inference and improves coverage and interpretive alignment. Sequential chunking alone reduces omission error by approximately 36–40\% and improves interpretive quality metrics by 20–50\%, with the largest proportional gains observed in 1B–3B models. This indicates that structured segmentation partially compensates for limited parameter capacity. However, sequential chunking still permits cumulative analytical bias, where early interpretations influence later synthesis and suppress less salient concepts.

\begin{wrapfigure}{l}{0.4\linewidth}
\vspace{-0.8cm}
    \centering
    \includegraphics[width=\linewidth]{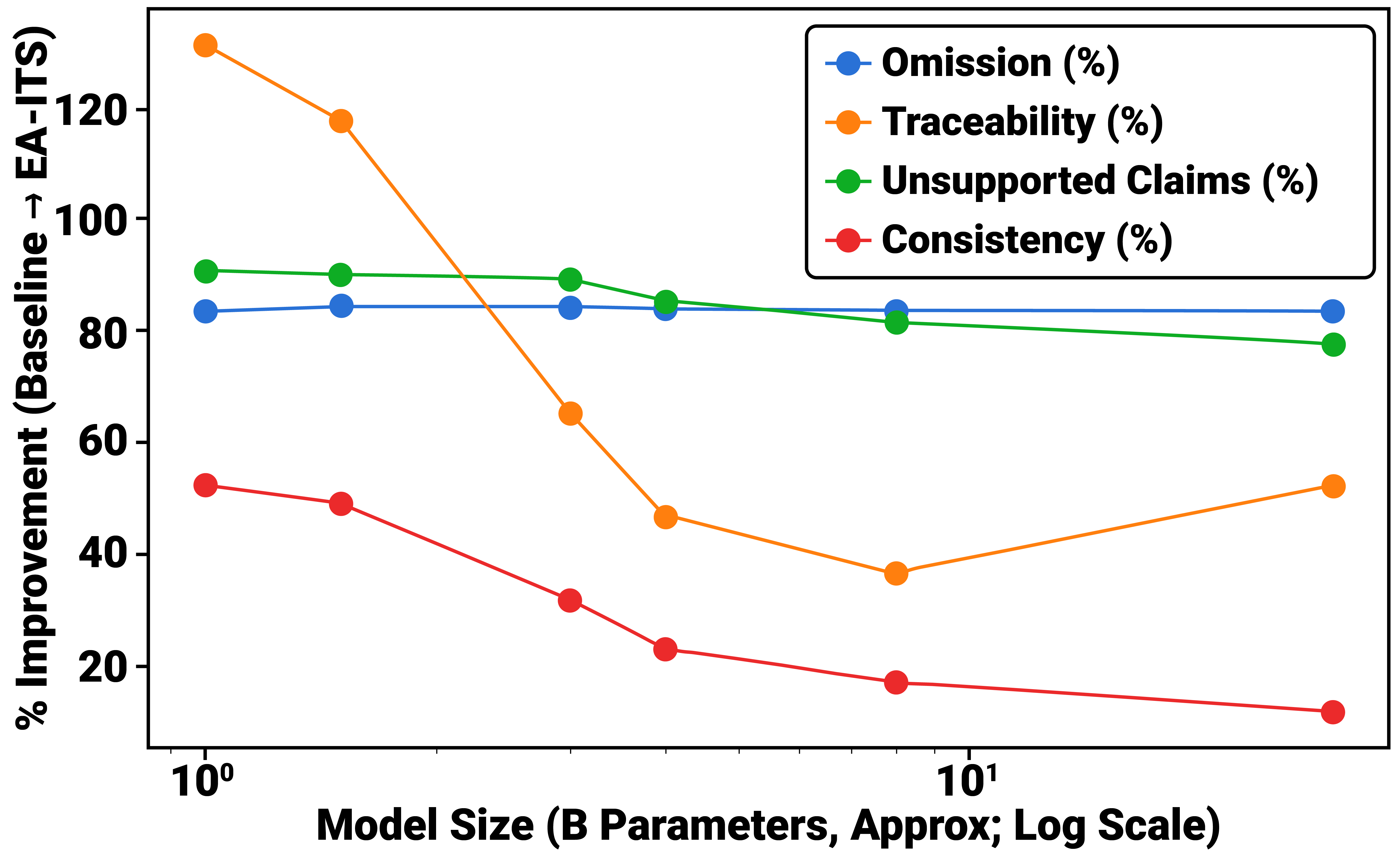}
    \caption{Scaling of PECII Performance Gains Across Model Sizes (Log-Scale Analysis of Improvement Trends)}
    \label{fig:scale}
    \vspace{-0.6cm}
\end{wrapfigure}

Parallel chunk-level execution resolves this limitation by ensuring independence across analytical units. Omission error decreases by approximately 80–84\% across all models, and early-chunk dominance is reduced by over 80\%, indicating that bias mitigation is primarily execution-driven rather than scale-driven. This pattern is reflected in Fig.~\ref{fig:scale}, where omission-related improvements remain consistently high across model sizes, confirming that parallel independence, not parameter growth, is the key determinant of bias reduction. Smaller models benefit proportionally more, narrowing the interpretive gap with larger architectures.

Beyond extraction, consolidation and prioritization are critical for interpretive rigor. Evidence-anchored synthesis significantly improves reliability by grounding interpretations in explicit textual support. Evidence traceability increases by 36–132\%, while unsupported claim rates decrease by approximately 78–91\%, directly reducing hallucination. Cross-theme leakage declines by 66–79\%, reflecting stronger conceptual boundary enforcement during merging. 

\begin{wrapfigure}{l}{0.65\linewidth}
\vspace{-0.8cm}
    \centering
    \includegraphics[width=\linewidth]{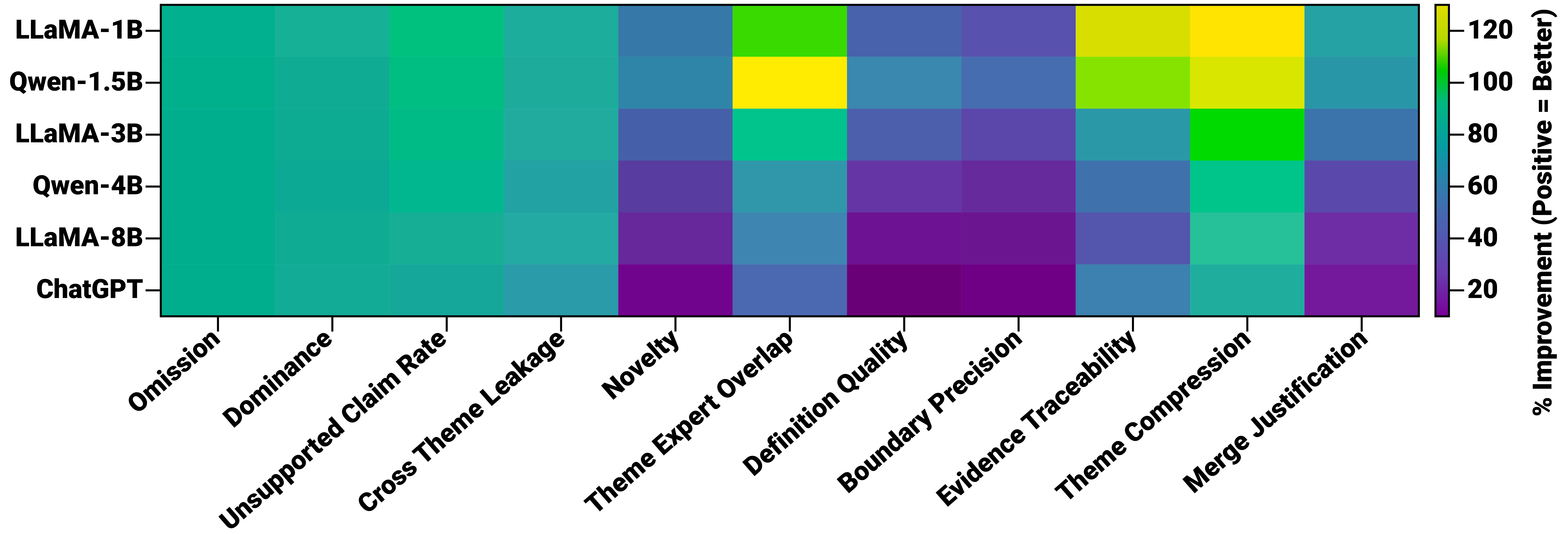}
    \caption{Global Improvement Heatmap: Percentage Change of Models and Evaluation Metrics}
    \label{fig:heatmap}
    \vspace{-0.7cm}
\end{wrapfigure}
The global improvement pattern is visible in Fig. \ref{fig:heatmap}, where structural and reliability-related metrics exhibit the strongest positive shifts across models. Notably, smaller models demonstrate larger proportional improvements, suggesting that evidence anchoring compensates for weaker intrinsic reasoning stability.
The findings indicate that reliable and scalable textual abstraction depends more on methodological architecture than on model scale alone. Chunking reduces interpretive overload, parallel execution eliminates cumulative dominance bias, and evidence-anchored consolidation suppresses unsupported claims while improving structural coherence, transforming both small and large models into more stable, traceable, and interpretable analytical systems.

\vspace{-0.5cm}

\section{Related Work}

LLMs based on transformer architecture have shown strong performance in text generation and reasoning \cite{raaian2024review,zhang2023ctg}. Research shows that increasing model size, training data, and compute improves performance \cite{shen2024efficient,menghani2023efficient}. LLMs also demonstrate few-shot and chain-of-thought reasoning abilities \cite{ma2023cot,nazi2025evaluation}. However, larger models still suffer from instability, hallucination, and sensitivity to prompt wording or order \cite{adeseye2026hallucination,rawte2023hallucination,li2023objecthallucination}. 
Processing long-form textual documents remains difficult due to limited context windows. Solutions include sparse attention \cite{yang2025lserve}, long-sequence transformers \cite{jacobs2024ulysses}, and retrieval-augmented models \cite{tural2024rag}. In practice, long-form textual data are often divided into chunks before analysis. While this improves efficiency, little research examines how chunk order may influence thematic interpretation. 
Autoregressive models depend on previous context, which can create anchoring effects \cite{campbell2009anchoring}. Early outputs may influence later reasoning, especially in multi-stage qualitative analysis. Although chain-of-thought improves reasoning, it can also propagate early errors \cite{chen2025reasoningera}. 
Hallucination is another major concern. LLMs sometimes generate unsupported or inaccurate claims \cite{latif2025hallucinations,banerjee2025hallucinate}. Retrieval augmentation and factual consistency methods are mitigation efforts, but most focus on single outputs rather than multi-stage thematic consolidation. 
Recent studies show that LLMs can support qualitative coding \cite{adeseye2025llmassisted,dai2023llmintheloop}, yet concerns remain about consistency, theme boundaries, and interpretive drift. Few studies analyze how different execution strategies (sequential vs. parallel chunking) affect bias, omission, or coherence. 
Overall, prior research studies focused on scaling, long-context handling, hallucination, and qualitative coding separately. However, limited work examines how execution strategy influences cumulative bias and evidence grounding in thematic abstraction. This study addresses that gap by proposing parallel chunk processing and evidence-constrained consolidation to improve reliability and reduce bias.

\vspace{-0.6cm}
\section{Conclusion}
\vspace{-0.3CM}
This study introduced PECII to improve the reliability of long-form document thematic abstraction with LLMs by addressing two structural failure modes: cumulative analytical bias from sequential chunk processing and ungrounded synthesis during consolidation. PECII combines boundary-aware chunking, order-invariant parallel inference, and evidence-anchored consolidation with trace pointers and semantic alignment checks. Across six models, results show that parallel execution substantially reduces omission error (80–84\%) and early-chunk dominance (80\%+), while improving interpretive novelty and expert-alignment metrics. Evidence anchoring further increases traceability (up to 130\%) and reduces unsupported claims (up to 90\%), while also lowering cross-theme leakage and improving merge justification quality. Importantly, structured execution narrows cross-model variance, suggesting that methodological inference design acts as a structural regularizer, enabling smaller models to approach large-model reliability ceilings. Overall, PECII advances LLM-based qualitative synthesis toward a more traceable, auditable, and reproducible pipeline. Future research should (i) test PECII across additional domains (e.g., policy, legal, clinical narratives) and multilingual corpora; (ii) refine evidence validation using entailment-based or verifier-model checks; (iii) study optimal chunking policies and adaptive thresholds for different transcript styles; (iv) evaluate human-in-the-loop variants where experts audit evidence links during consolidation; and (v) quantify computational trade-offs and parallelization strategies for resource-constrained deployments. 
\vspace{-0.3CM}
\section{Declaration on the Use of Generative AI}
Language editing and grammar-checking tools were used to improve clarity and readability of the manuscript.

\section{Appendix}

\subsection{Detailed Sequential and Consolidation Modeling}

Sequential processing makes $\widehat{\Omega}$ depend on the order in which chunks are processed. Let $\pi$ be a permutation over the $M$ chunks. Sequential inference processes chunks as:

{\scriptsize
\begin{equation}
c_{\pi(1)},c_{\pi(2)},\dots,c_{\pi(M)}
\end{equation}
}

Let $\mathcal{F}_\theta$ denote the model-based inference operator with parameters $\theta$. Let $s_t$ denote the evolving analysis state after processing the first $t$ chunks. Then:

{\scriptsize
\begin{equation}
\widehat{\Omega}_t = \mathcal{F}_\theta(c_{\pi(t)}; s_t),
\qquad
s_t = \Psi(s_{t-1},\widehat{\Omega}_{t-1})
\end{equation}
}

The final extracted set under sequential processing is:

{\scriptsize
\begin{equation}
\widehat{\Omega}_{\text{seq}}(\pi)=\bigcup_{t=1}^{M}\widehat{\Omega}_t
\end{equation}
}

We model reinforcement using a dominance distribution:

{\scriptsize
\begin{equation}
P_t(\omega)=
\frac{
\exp\left(
\lambda \sum_{i=1}^{t}\alpha_{t,i}\,\mathbb{I}[\omega\in \widehat{\Omega}_i]
\right)
}{
\sum_{\omega'\in\Omega}
\exp\left(
\lambda \sum_{i=1}^{t}\alpha_{t,i}\,\mathbb{I}[\omega'\in \widehat{\Omega}_i]
\right)
}
\end{equation}
}

Extraction probability:

{\scriptsize
\begin{equation}
\Pr(\omega \in \widehat{\Omega}_t \mid c_{\pi(t)}, s_t)
=
\sigma\left(
A_\theta(\omega,c_{\pi(t)})
-
\eta P_{t-1}(\omega)
-
\zeta\sum_{\omega'\neq \omega}P_{t-1}(\omega')\Gamma(\omega,\omega')
\right)
\end{equation}
}

Expected omission:

{\scriptsize
\begin{equation}
\mathbb{E}[\mathcal{E}_{\text{om}}(\widehat{\Omega}_{\text{seq}})]
=
1-\frac{1}{|\Omega^\star|}
\sum_{\omega\in\Omega^\star}
\Pr\left(\exists t: \omega\in \widehat{\Omega}_t\right)
\end{equation}
}

Semantic alignment:

{\scriptsize
\begin{equation}
\mathcal{S}(\omega,e)=
\cos\left(\mathrm{emb}(h(\omega)),\mathrm{emb}(q(e))\right)
\end{equation}
}

Similarity graph and clustering:

{\scriptsize
\begin{equation}
G = (V,E),
\quad
E = \{(i,j): \cos(v_i,v_j) \ge \delta\}.
\end{equation}
}

{\scriptsize
\begin{equation}
\mathcal{R}(\mathcal{K})
=
\sum_{k}
\sum_{\substack{i<j\\ z_i,z_j\in K_k}}
\cos(v_i,v_j)
\end{equation}
}

{\scriptsize
\begin{equation}
\mathcal{L}(\mathcal{K})
=
\sum_{k\ne \ell}
\frac{|E(K_k)\cap E(K_\ell)|}{|E(K_k)\cup E(K_\ell)|}
\end{equation}
}

Order sensitivity estimator:

{\scriptsize
\begin{equation}
\mathcal{B}_{\text{order}}
=
\mathbb{E}_{\pi}\Big[
d\big(\widehat{\Omega}_{\text{seq}}(\pi), \mathbb{E}_{\pi'}[\widehat{\Omega}_{\text{seq}}(\pi')]\big)
\Big]
\end{equation}
}

\subsection{Additional Analysis}

\paragraph{Formula used}

\textbf{Theme--Expert Overlap (TEO):}
This measures alignment between model-generated themes and independent expert themes:
\vspace{-0.15cm}
{\small
\begin{equation}
\mathrm{TEO} =
\frac{|T_{\text{model}} \cap T_{\text{expert}}|}
{|T_{\text{expert}}|}
\vspace{-0.15cm}
\end{equation}
}
It assesses interpretive validity.  
Replication requires independent expert coding, semantic theme matching via cosine similarity (threshold $\geq 0.80$), and computation of overlap ratio.

\textbf{Definition Quality Score (1--5):}
Assesses clarity and coherence of theme definitions:
\vspace{-0.15cm}
{\small
\begin{equation}
\mathrm{DQS} =
\frac{1}{|\Theta|}
\sum_{\theta \in \Theta} r(\theta)
\vspace{-0.15cm}
\end{equation}
}
This prevents vague or overgeneralized abstractions.  
Replication requires blind expert ratings and averaging across raters, optionally reporting inter-rater agreement.

\textbf{Boundary Precision Score (\%):}
Measures accuracy of excerpt inclusion and exclusion decisions:
\vspace{-0.15cm}
{\small
\begin{equation}
\mathrm{BPS} =
\frac{\text{Correct Decisions}}
{\text{Total Decisions}}
\times 100
\end{equation}
}

It enforces conceptual boundary control and prevents cross-theme leakage.  
Replication requires sampling excerpt-theme assignments, expert validation, and computing percentage correct.

\paragraph{Results}

The complete information about the results can be seen in Table \ref{tab:ea_its_quality_colored_updated}.

\begin{table*}[htbp]
\centering
\scriptsize
\vspace{-0.3cm}
\caption{Interpretive Quality Evaluation Across Models Under Different Execution Strategies}
\label{tab:ea_its_quality_colored_updated}
\renewcommand{\arraystretch}{1.2}
\setlength{\tabcolsep}{1.1pt}

\newcommand{\chgcellbaseline}[5]{%
\begingroup
\setlength{\fboxsep}{0.9pt}
\setlength{\fboxrule}{0.6pt}
\begin{tabular}[c]{@{}c@{}}%
{\scriptsize
\fcolorbox{orange!60!black}{orange!15}{\textbf{#4}}%
\hspace{3pt}%
\fcolorbox{green!60!black}{green!15}{\textbf{#5}}}\\[-0.5pt]
\textcolor{blue}{#1} $\rightarrow$ \textcolor{orange}{#2} / \textcolor{red}{#3}
\end{tabular}%
\endgroup
}

\begin{tabular}{|l|c|c|c|}
\hline
\textbf{Model} 
& \textbf{\makecell{Theme--Expert Overlap}} 
& \textbf{\makecell{Definition Quality}} 
& \textbf{\makecell{Boundary Precision(\%)}} \\
\hline

LLaMA-1B &
\chgcellbaseline{0.42}{0.63}{0.89}{+50.0\%}{+112.0\%} &
\chgcellbaseline{3.4}{4.1}{4.8}{+20.6\%}{+41.2\%} &
\chgcellbaseline{70}{84}{94}{+20.0\%}{+34.3\%} \\
\hline

Qwen-1.5B &
\chgcellbaseline{0.38}{0.60}{0.90}{+57.9\%}{+136.8\%} &
\chgcellbaseline{3.1}{4.0}{4.8}{+29.0\%}{+54.8\%} &
\chgcellbaseline{66}{85}{95}{+28.8\%}{+43.9\%} \\
\hline

LLaMA-3B &
\chgcellbaseline{0.48}{0.69}{0.91}{+43.8\%}{+89.6\%} &
\chgcellbaseline{3.5}{4.2}{4.9}{+20.0\%}{+40.0\%} &
\chgcellbaseline{72}{87}{95}{+20.8\%}{+31.9\%} \\
\hline

Qwen-4B &
\chgcellbaseline{0.56}{0.74}{0.92}{+32.1\%}{+64.3\%} &
\chgcellbaseline{3.9}{4.4}{4.9}{+12.8\%}{+25.6\%} &
\chgcellbaseline{78}{90}{96}{+15.4\%}{+23.1\%} \\
\hline

LLaMA-8B &
\chgcellbaseline{0.60}{0.76}{0.92}{+26.7\%}{+53.3\%} &
\chgcellbaseline{4.2}{4.6}{4.9}{+9.5\%}{+16.7\%} &
\chgcellbaseline{82}{92}{96}{+12.2\%}{+17.1\%} \\
\hline

ChatGPT &
\chgcellbaseline{0.64}{0.78}{0.91}{+21.9\%}{+42.2\%} &
\chgcellbaseline{4.5}{4.7}{4.9}{+4.4\%}{+8.9\%} &
\chgcellbaseline{85}{92}{95}{+8.2\%}{+11.8\%} \\
\hline

\end{tabular}
\vspace{-0.4cm}
\end{table*}
\FloatBarrier


Table \ref{tab:ea_its_quality_colored_updated} shows consistent improvements in interpretive rigor across all models. At baseline (no chunking), performance generally increases with model size. For example, Theme–Expert Overlap rises from 0.42 in LLaMA-1B to 0.64 in ChatGPT. Although small minor 
fluctuations appear among smaller models (for example, Qwen-1.5B is slightly below LLaMA-1B). The overall pattern reflects improvement with increasing model size (i.e., parameter count) under sequential execution. Therefore, larger models demonstrated stronger interpretive alignment, clearer definitions, and better boundary control when processing transcripts sequentially. Sequential chunking substantially improves performance across all metrics. Theme–Expert Overlap increases by approximately 21.9–57.9\% relative to baseline. 
Parallel chunking produces even larger gains (42.2–136.8\%), with the largest improvements observed in smaller models (e.g., Qwen-1.5B +136.8\%), indicating that execution strategy can strongly compensate for lower baseline Theme–Expert Overlap performance.
However, this pattern partly reflects denominator effects: models with lower starting baseline scores naturally show larger relative gains. In absolute terms, both small and large models improve meaningfully. Relative improvements decrease smoothly as baseline performance increases, indicating a clear diminishing-return pattern.


Definition Quality and Boundary Precision follow the same structural pattern. Sequential chunking improves clarity and boundary control by approximately 4.4–29\%, while parallel chunking improves them by 8.9–54.8\%, depending on model size. Smaller models showed greater relative improvements, while larger models (e.g., ChatGPT) experienced comparatively smaller gains, likely due to their stronger baseline performance and reduced potential for additional improvement due to stronger reasoning capability. With parallel execution, differences across model scales decrease substantially, indicating that execution structure plays a major role in determining final interpretive quality.
Across all metrics, overlap, definition quality, and boundary precision increase under both chunking strategies, with no trade-offs observed. This consistent improvement suggests that the gains arise from changes in the inference structure rather than from properties of a specific evaluation metric. A likely explanation relates to autoregressive conditioning. In sequential chunking, early themes influence later interpretations, reinforcing dominant concepts and creating cumulative anchoring effects. Each output becomes part of the next input, leading to the development of semantic reinforcement loops. Parallel chunking interrupts this recursive dependency by isolating chunk interpretations and resetting the conditioning state for each segment. By preventing cumulative anchoring and cross-chunk interference, parallel chunking reduces scale-dependent bias accumulation and narrows model architectural differences while simultaneously improving overall interpretive quality. Generally, while model architectural type and parameter size determine baseline performance, structural inference strategy, particularly parallel chunk-level execution, substantially reduces cross-model disparities and improves performance for all models.

%
%

\end{document}